\pdfoutput=1

\documentclass[11pt]{article}

\usepackage[]{ACL2023}

\usepackage{times}
\usepackage{graphicx}
\usepackage{latexsym}

\usepackage[T1]{fontenc}

\usepackage[utf8]{inputenc}

\usepackage{microtype}

\usepackage{inconsolata}

\title{Knowledge of cultural moral norms in large language models}

\author{Aida Ramezani \\
  Department of Computer Science \\University of Toronto \\
  \texttt{armzn@cs.toronto.edu} \\\And
  Yang Xu \\
  Department of Computer Science \\
  Cognitive Science Program \\
  University of Toronto  \\
  \texttt{yangxu@cs.toronto.edu} \\}
  
\begin{document}
\maketitle
\begin{abstract}
Moral norms vary across cultures. A recent line of work suggests that English large language models contain human-like moral biases, but these studies typically do not examine moral variation in a diverse cultural setting. We investigate the extent to which monolingual English language models contain knowledge about moral norms in different countries. We consider two levels of analysis: 1) whether language models capture fine-grained moral variation across countries over a variety of topics 
such as ``homosexuality'' and ``divorce''; 2) whether language models capture cultural diversity and shared tendencies in which topics people around the globe tend to diverge or agree on in their moral judgment. We perform our analyses with two public datasets from the World Values Survey (across 55 countries) and PEW global surveys (across 40 countries) on morality. We find that pre-trained English language models predict empirical moral norms across countries worse than the English moral norms reported previously. However, fine-tuning language models on the survey data improves inference 
across countries at the expense of a less accurate estimate of the English moral norms. We discuss the relevance and challenges of incorporating cultural knowledge into the automated inference of moral norms. 
 
\end{abstract}

\section{Introduction}
\begin{figure*}
\includegraphics[width=\textwidth]{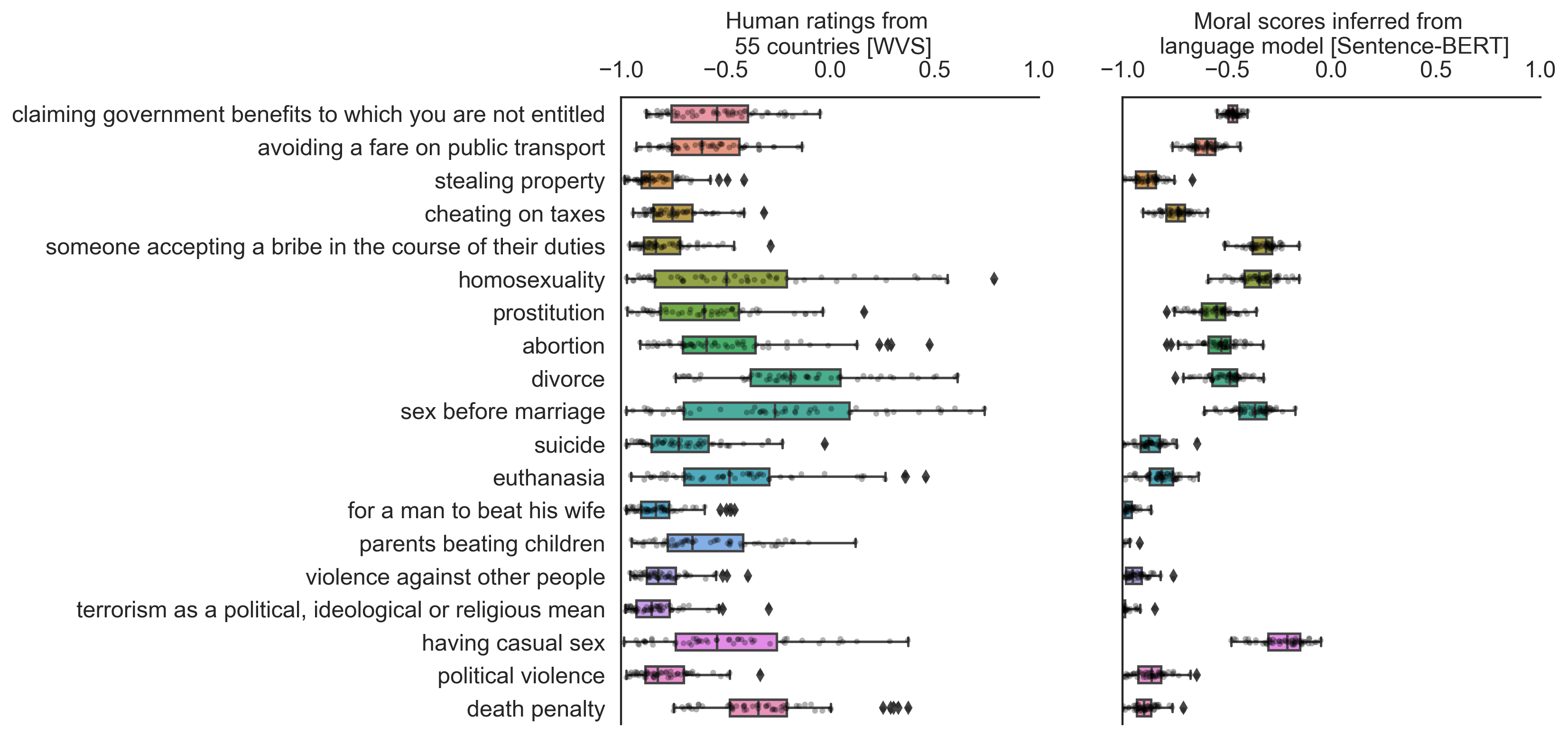}
  \caption{Comparison of human-rated and machine-scored moral norms across cultures. Left: Boxplots of human ratings of moral norms across countries in the World Values Survey (WVS)~\cite{haerpfer2021world}. Each dot represents the empirical average of participants' ratings for a morally relevant topic (e.g., ``abortion'') within a country. Right: Corresponding moral scores estimated by a language model (Sentence-BERT)~\cite{reimers-2019-sentence-bert}. Each dot represents the moral score obtained by probing the language model in a given country.}
\label{fig:wvs_fig1}
\end{figure*}

Moral norms vary from culture to culture~\cite{haidt1993affect,bicchieri2005grammar,atari22morality,iurino2020testing}. Understanding the cultural variation in moral norms has become critically relevant to the development of machine intelligence. For instance, recent work has shown that cultures vary substantially in their judgment toward moral dilemmas regarding autonomous driving~\cite{awad2018moral, awad2020universals}. 
Work in Natural Language Processing (NLP) also shows that language models capture some knowledge of social or moral norms and values. For example, with no supervision, English pre-trained language models (EPLMs) have been shown to capture people's moral biases  and distinguish between morally right and wrong actions~\cite{schramowski2022large}. Here we investigate whether EPLMs encode knowledge about moral norms across cultures, an open issue that has not been examined comprehensively.

Multilingual pre-trained language models (mPLMs) have been probed for their ability to identify cultural norms and biases  in a restricted setting~\cite{yin2022geomlama, arora2022probing, hammerl2022multilingual, touileb-etal-2022-occupational}. For instance,~\citet{hammerl2022multilingual} show that mPLMs capture moral norms in a handful of cultures that speak different languages. However, it remains unclear whether monolingual EPLMs encode cultural knowledge about moral norms. Prior studies have only used EPLMs to assess how they encode undesirable biases toward different communities~\cite{ousidhoum-etal-2021-probing,abid2021persistent, sap-etal-2020-social,nozza-etal-2021-honest,nozza-etal-2022-measuring}. For instance,~\citet{abid2021persistent} show that \textsc{GPT3} can generate toxic comments against Muslims, and~\citet{nozza-etal-2022-measuring} explore  harmful text generation toward LGBTQIA+ groups in BERT models ~\cite{devlin2018bert,liu2019roberta}.

Extending these lines of work, we assess whether monolingual EPLMs can accurately infer moral norms across many cultures. Our focus on EPLMs is due partly to the fact that English as a lingua franca has widespread uses for communication in-person and through online media. Given that EPLMs may be applied to  multicultural settings, it is important to understand whether these models encode basic knowledge about cultural diversity. Such knowledge  has both relevance and applications for NLP such as automated toxicity reduction and content moderation~\cite{schramowski2022large}. Another motivation for our focus is that while it is expected that EPLMs should encode western and English-based moral knowledge, such 
knowledge might entail potential (implicit) biases toward non-English speaking cultures. For example, an EPLM might infer a situation to be morally justifiable (e.g., ``political violence'') in a non-English speaking culture (because these events tend to associate with non-English speaking cultures in corpora) and thus generate misleading representations of that community. 

Here we probe state-of-the-art EPLMs trained on large English-based datasets. Using EPLMs also supports
a scalable analysis of 55 countries, which goes beyond existing work focusing on a small set of high-resource languages from mPLMs and monolingual PLMs. We take the moral norms reported in different countries to be a proxy of cultural moral norms and consider two main levels of analysis to address the following questions:
\begin{itemize}
    \item Level 1: Do EPLMs encode moral knowledge that mirrors the moral norms in different countries? For example, ``getting a
divorce'' can be a morally frowned-upon topic in country
$i$, but morally acceptable in country $j$.  
\item Level 2: Can EPLMs infer the cultural diversity and shared tendencies in moral judgment of different topics? For example, people across nations might agree that doing $X$ is morally wrong while disagreeing in their moral judgment toward $Y$. 
\end{itemize}

We probe EPLMs using two publicly available global surveys of morality, World Values Survey wave 7~\cite{haerpfer2021world}~\footnote{\url{https://www.worldvaluessurvey.org/WVSContents.jsp}} (WVS) and PEW Global Attitudes survey (PEW)~\cite{pew}~\footnote{\url{https://www.pewresearch.org/global/interactives/global-morality/}}. 
For example, according to WVS survey (illustrated in Figure~\ref{fig:wvs_fig1}), people in different cultures hold disparate views on whether ``having casual sex'' is morally acceptable. In contrast, they tend to agree more about the immorality of ``violence against other people''. Our level 1 analysis allows us to probe the fine-grained cultural moral knowledge in EPLMs, and our level 2 analysis investigates the EPLMs' knowledge about shared ``universals'' and variability across cultures in moral judgment. 
Following previous work~\cite{arora2022probing} and considering the current scale of global moral surveys, we use country as a proxy to culture, although this approach is not fully representative of all the different cultures within a country.

We also explore the utility-bias trade-off in encoding the knowledge of cultural moral norms in EPLMs through a fine-tuning approach. With this approach it may be possible to enhance the moral knowledge of EPLMs in a multicultural setting. We examine how this approach might reduce the ability of EPLMs to infer English-based moral norms and discuss how it might induce cultural biases.


\section{Related work}

\subsection{Automated moral inference in  NLP}

Large language models have been utilized to make automated moral inference from text.~\citet{trager2022moral} used an annotated dataset to fine-tune language models to predict the moral foundations~\cite{graham2013moral} expressed in Reddit comments. Many other textual datasets and methods have been proposed for fine-tuning LMs for moral norm generation, reasoning, and adaptation~\cite{forbes2020social,emelin-etal-2021-moral, hendrycks2021aligning, ammanabrolu-etal-2022-aligning, liu2022aligning, lourie2021scruples,Jiang2021DelphiTM}.
~\citet{schramowski2022large} proposed a method to estimate moral values and found EPLMs to capture human-like moral judgment even without fine-tuning. They identified a \textsc{MoralDirection} using the semantic space of Sentence-BERT~\cite{reimers-2019-sentence-bert} (\textsc{SBERT}) that corresponds to values of right and wrong. The semantic representations of different actions (e.g., \textit{killing people}) would then be projected in this direction for moral judgment estimation. However, this method assumed a homogeneous set of moral norms, so it did not examine cultural diversity in moral norms.

\subsection{Language model probing}

Probing has been used to study knowledge captured in language models. ~\citet{petroni-etal-2019-language} proposed a methodology to explore the factual information that language models store in their weights. Similar probing techniques have been proposed to identify  harmful biases captured by PLMs.~\citet{ousidhoum-etal-2021-probing} probed PLMs to identify toxic contents that they generate toward people of different communities.~\citet{nadeem-etal-2021-stereoset} took a similar approach and introduced Context Association Tests to measure the stereotypical biases in PLMs,~\citet{yin2022geomlama} used probing to evaluate mPLMs on geo-diverse commonsense knowledge, and ~\citet{touileb-etal-2022-occupational} developed probing templates to investigate the occupational gender biases in multilingual and Norwegian language models.
Related to our work,~\citet{arora2022probing} used cross-cultural surveys to generate prompts for evaluating mPLMs in 13 languages. For each country and category (e.g., Ethical Values) in the surveys, they take an average of participants' responses to different questions in the category and show that mPLMs do not correlate with the cultural values of the countries speaking these languages. Differing from that study, we assess finer-grained prediction of EPLMs on people's responses to individual survey questions. More recently,~\citet{DILLION2023} 
prompted GPT-3.5~\cite{brown2020language} 
with human judgments in different moral scenarios and found striking correlation between the model outputs and the human judgments. Similar to~\citet{schramowski2022large}, this work also used a homogeneous set of moral ratings which represented English-based and Western cultures.

\section{Methodology for inferring cultural moral norms}
\label{sec:methods}

We develop a method for fine-grained moral norm inference across cultures. This method allows us to probe EPLMs with topic-country pairs, such as ``getting a divorce in [Country]''.\footnote{We replace [Country] with a country's name.} We build this method from the baseline method proposed by~\citet{schramowski2022large} for homogeneous moral inference, where we probe EPLM's moral knowledge about a topic without incorporating the cultural factor (i.e., the country names). Similar to that work, we use \textsc{SBERT} through \texttt{bert-large-nli-mean-tokens} sentence transformer model and use topic and topic-country pairs as our prompts.\footnote{We make our code and data available on \url{https://github.com/AidaRamezani/cultural\_inference}.} This model is built on top of the \textsc{BERT} model, which is pre-trained on \textsc{BooksCorpus}~\cite{DBLP:journals/corr/ZhuKZSUTF15} and Wikipedia.




\subsection{Autoregressive EPLMs}
Since the \textsc{MoralDirection} is constructed from the semantic space of the \textsc{BERT}-based EPLMs~\cite{schramowski2022large}, we develop a novel approach to probe autoregressive state-of-the-art EPLMs,  GPT2~\cite{radford2019language} and GPT3~\cite{brown2020language}.
For each topic or topic-country pair, we construct the input $s$ as ``In [Country] [Topic]''. We then append a pair of opposing moral judgments to $s$ and represent them formally as $(s^{+}, s^{-})$. For example, for $s = $ ``In [Country] getting a divorce'', and (\textit{always justifiable, never justifiable}) as the moral judgment pair, $s^{+}$ and $s^{-}$ would be ``In [Country] getting a divorce is always justifiable'' and ``In [Country] getting a divorce is never justifiable'' respectively.\footnote{We also try probing with the template $s =$ ``People in [Country] believe [Topic]'', but the results do not improve, so we report the most optimal prompts in the main text, and the rest are shown in Appendix~\ref{sec:prompts_details}.} To make our probing robust to the choice of moral judgments, we use a set of $K = 5$ prompt pairs (i.e.,\{\textit{(always justifiable, never justifiable), (morally good, morally bad), (right, wrong), (ethically right, ethically wrong), (ethical, unethical)}\}), and refer to appended input pairs as $(s_i^{+}, s_i^{-})$ where $i \in [K]$.
Since \textsc{GPT2} and \textsc{GPT3} are composed of decoder blocks in the transformer architecture~\cite{vaswani2017attention}, we use the probabilities of the last token in $s_i^{+}$, and $s_i^{-}$ as a moral score for each. The moral score of the pair $(s_i^{+}, s_i^{-})$ is the difference between the log probabilities of its positive and negative statements. 
\begin{equation}
    MS(s_{i}^{+}, s_{i}^{-}) = \log \frac{P(s_{i T}^{+} | s_{i < T}^{+})} {P(s_{i T}^{-} | s_{i < T}^{-})}
\label{eq:moral_value}
\end{equation}
Here $s_{i T}^{+}$ and $s_{i T}^{-}$ are the last tokens in $s_i^{+}$ and $s_i^{-}$ respectively, and their probabilities can be estimated by the softmax layer in autoregressive EPLMs. 

We take an average of the estimated moral scores for all $K$ pair statements to compute the moral score of the input. 

\begin{equation}
    MS(s) = \frac{1}{K} \sum_{i = 1}^{K} MS(s_i^{+}, s_i^{-})
\end{equation}
To construct the baseline, we compute the homogeneous moral score of a topic without specifying the country in the prompts. Using prompt pairs allows us to operationalize moral polarity: a positive moral score indicates that on average the EPLM is more likely to generate positive moral judgment for  input $s$, compared to negative moral judgment.

We use \textsc{GPT2} (117M parameters), \textsc{GPT2-MEDIUM} (345M parameters), \textsc{GPT2-LARGE} (774M parameters), and \textsc{GPT3} (denoted as \textsc{GPT3-PROBS}, 175B parameters)\footnote{We access \textsc{GPT2} through \texttt{transformer} package provided by \texttt{Huggingface}.
We access \textsc{GPT3} through OpenAI API of \texttt{text-davinci-002} engine with a temperature of $0.6$ for text generation.}. 
\textsc{GPT2} is trained on \textsc{WebText}, which is a dataset of webpages and contains very few non-English samples. Around $82\%$ of the pre-training data for \textsc{GPT3} comes from Common Crawl data and \textsc{WebText2}~\cite{kaplan2020scaling}, an extended version of \textsc{WebText}~\cite{radford2019language}. Around $7\%$ of the training corpus of \textsc{GPT3} is non-English text. Considering such data shift from books and articles in \textsc{BERT} to webpages in \textsc{GPT2} and \textsc{GPT3} in astronomical sizes, it is interesting to observe how cultural moral norms would be captured by EPLMs trained on webpages, which cover a more heterogeneous set of contents and authors.

We also design multiple-choice question prompts to leverage the question-answering capabilities of \textsc{GPT3} (denoted as \textsc{GPT3-QA}). Similar to the wording used in our ground-truth survey datasets, questions are followed by three options each describing a degree of moral acceptability. We repeat this question-answering process $5$ times for each topic-country pair and take the average of the model responses. Table~\ref{tab:prompt} in the Appendix shows our prompts for all models.

\section{Datasets}
We describe two open survey data that record moral norms across cultures over a variety of topics.

\subsection{World Values Survey}
The Ethical Values section in World Values Survey Wave 7 (WVS for short) is our primary dataset. This wave covers the span of 2017-2021 and is publicly available~\cite{haerpfer2021world}. 
In the Ethical Values section, participants from 55 countries were surveyed regarding their opinions on 19 morally-related topics. The questionnaire was translated into the first languages spoken in each country and had multiple options. We normalized the options to range from $-$1 to 1, with $-$1 representing ``never justifiable'' and 1 ``always justifiable''. The moral rating of each country on each topic (i.e., topic-country pair) would then be the average of the participant's responses.

\subsection{PEW 2013 global attitude survey}
We use a secondary dataset from PEW Research Center~\cite{pew} based on a public survey in 2013 that studied global moral attitudes in 40 countries toward eight morally-related topics (PEW for short). 100 people from each country participated in the survey. The questions were asked in English and had three options representing ``morally acceptable'', ``not a moral issue'', and ``morally unacceptable''. We normalized these ratings to be in the range of $-$1 to 1 and represented each topic-country pair by taking an expected value of all the responses.

\subsection{Homogeneous moral norms}
We also use the data from the global user study in~\citet{schramowski2022large} which were collected via Amazon MTurk from English speakers. This dataset contains $234$ participants' aggregated ratings of moral norms used for identifying the \textsc{MoralDirection}. Around half of the participants are from North America and Europe. We refer to this dataset as ``Homogeneous norms'' since it does not contain information about moral norms across cultures.

\section{Evaluation and results}
We evaluate EPLMs' moral knowledge with respect to 1) homogeneous moral norms, 2) fine-grained moral norms across cultures, and 3) cultural diversities and shared tendencies on  moral judgment of different topics.

\subsection{Homogeneous moral norm inference}
For homogeneous moral norm inference, we compute Pearson correlation between 1) the empirical homogeneous moral ratings, obtained by aggregating the human moral ratings toward a topic from all countries, and 2) language model inferred moral scores, estimated from  our homogeneous probing method (i.e., without specifying country in prompts). 

\begin{figure} 
\centering
\includegraphics[width=0.47\textwidth]{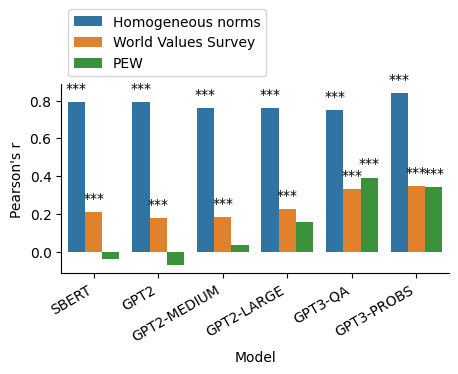}
\caption{Performance of EPLMs (without cultural prompts) on inferring 1) English  moral norms, and 2) culturally diverse moral norms recorded in World Values Survey and PEW survey data. The asterisks indicate the significance levels (``*'', ``**'', ``***'' for $p<0.05, 0.01, 0.001$ respectively). }
  \label{fig:tables_merged}
\end{figure}

 Figure~\ref{fig:tables_merged} shows the results on World Values Survey ($n = 1,028$), PEW survey ($n = 312$), and the Homogeneous norms datasets ($n = 100$). The high correlation of \textsc{GPT2} and \textsc{GPT3} moral scores with the Homogeneous norms dataset indicate that our methodology does indeed capture the embedded moral biases in these models, with similar performance to the method proposed by~\citet{schramowski2022large} for \textsc{SBERT} ($r = 0.79$), and higher for \textsc{GPT3-PROBS} ($r = 0.85$). The moral norms in this dataset are typically more globally agreeable (e.g., \textit{You should not kill people}) than topics in WVS and PEW. As expected, EPLMs are less correlated with WVS and PEW, since their moral biases are derived from pre-training on English and westernized data. Aggregated ratings in WVS and PEW, however, capture a more global view toward moral issues, which are also morally contentious (e.g., ``getting a divorce''). Table~\ref{tab:varrying_corr} in Appendix includes the values for this experiment.


\subsection{Fine-grained cultural variation of moral norms toward different topics}
\label{sec:cultural_variation}
Going beyond probing EPLMs for their general knowledge of moral norms, we assess whether they can accurately identify the moral norms of different cultures (level 1 analysis). Using our fine-grained probing approach described in Section~\ref{sec:methods}, we compute Pearson correlation between EPLMs' moral scores and the fine-grained moral ratings from the ground truth. Each sample pair in the correlation test corresponds to 1) the moral norms estimated by EPLMs for a country $c$ and a topic $t$, and 2) the empirical average of moral ratings toward topic $t$ from all the participants in the country $c$.

\begin{figure*}[ht!]
\centering
\includegraphics[width=\textwidth]{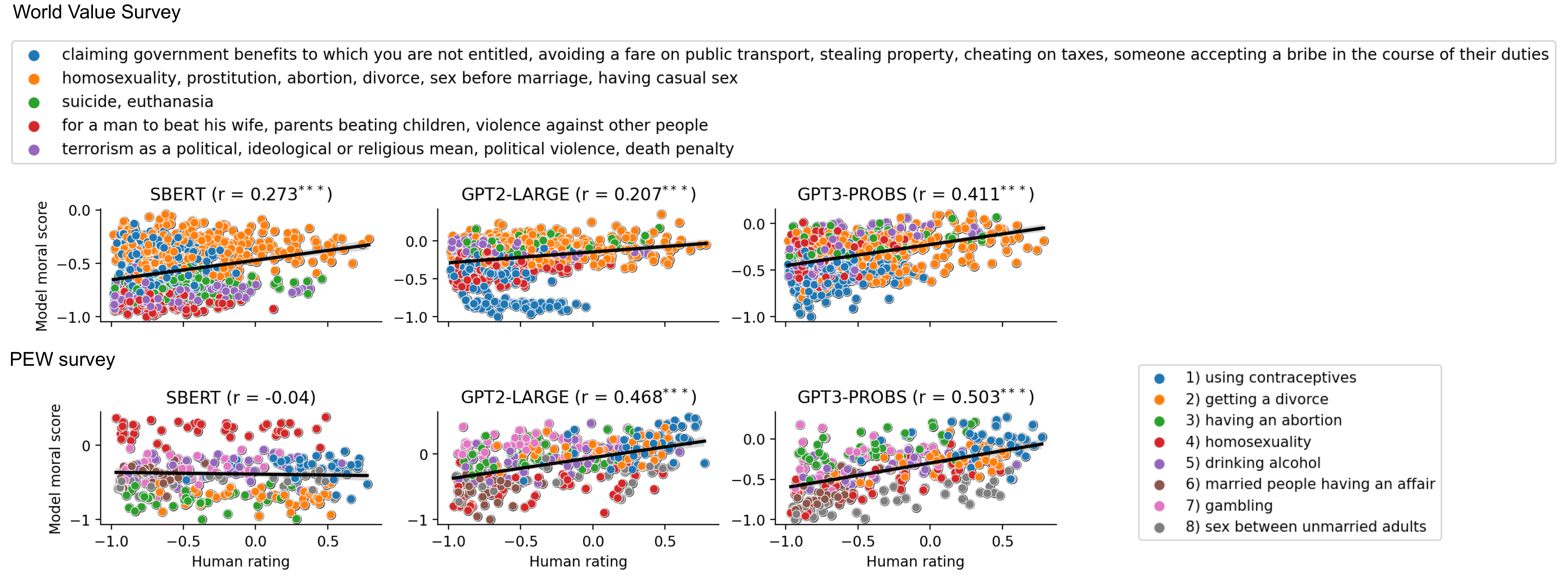}
\caption{Degree of alignment between the moral scores from EPLMs and fine-grained empirical moral ratings for different topics across countries taken from the World Values Survey (top) and PEW survey (bottom). 
Each dot represents a topic-country pair. The x-axis shows the fine-grained moral ratings from the ground truth and the y-axis shows the corresponding inferred moral scores. The legends display the moral topics in the surveys. Similar topics in the World Value Surveys are shown with the same color. }
\label{fig:figure2}
\end{figure*}

Figure~\ref{fig:figure2} summarizes the results for \textsc{SBERT, GPT2-LARGE}, and \textsc{GPT3-PROBS} models, and the rest of the models are shown in Figure~\ref{fig:figure2_app} in the Appendix. 
To facilitate direct comparison, the estimated moral scores are normalized to a range of $-1$ to $1$, where $-$1, 0, and 1 indicate morally negative, morally neutral, and morally positive norms, respectively. 
\textsc{GPT3-QA} and \textsc{GPT3-PROBS} both show a relatively high correlation with the cultural variations of moral norms ($r = 0.352$, $r = 0.411$, $p < 0.001$, for both), and \textsc{GPT2-LARGE} achieves a correlation of $r = 0.207$ ($p < 0.001$) in WVS where $n = 1,028$. The correlations are relatively better for PEW ($n = 312$) with $r = 0.657$, $r = 0.503$, and $r = 0.468$ for \textsc{GPT3-QA, GPT3-PROBS} and \textsc{GPT2-LARGE} respectively. These results show that EPLMs have captured some knowledge about the moral norms of different cultures, but with much less accuracy (especially for \textsc{GPT2} and \textsc{SBERT}) compared to their inference of English moral norms shown in the previous analysis. 

In addition, we check whether \textsc{GPT3}'s high correlation with PEW is because it has seen and memorized the empirical data. Our investigation shows that \textsc{GPT3} has seen the data during pre-training, as it can generate the sentences used on the survey website. However, the scores suggested by \textsc{GPT3} text generation and the countries' rankings based on their ratings are different from the ground truth data. 
 
\subsection{Culture clustering through fine-grained moral inference}
\label{sec:clusters}
\begin{figure*}
\includegraphics[width=\textwidth]{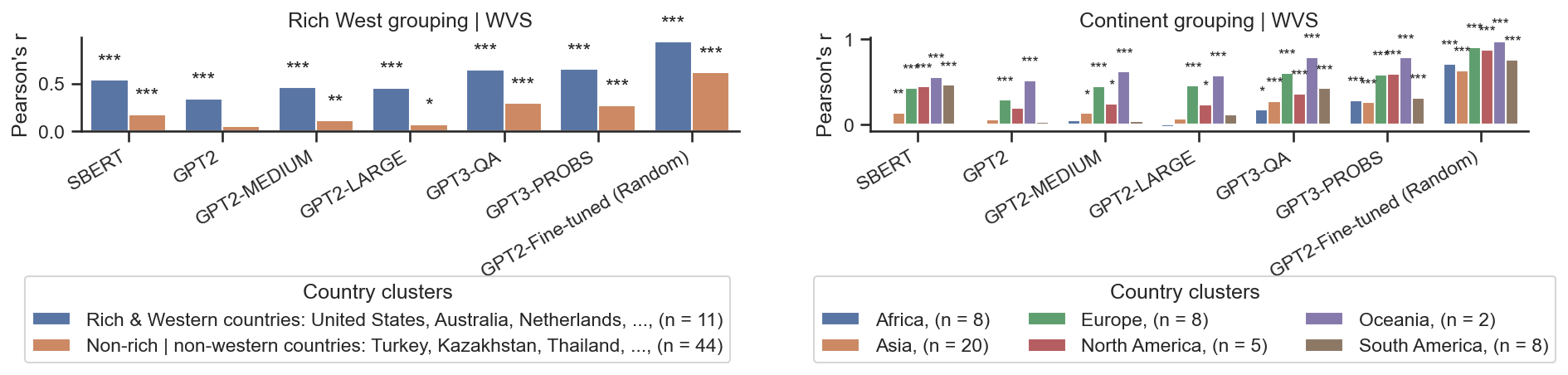}
\caption{Correlation between language-model inferred moral scores and empirical moral ratings from World Values Survey, analyzed in different clusters of countries in Rich West grouping (left) and continent grouping (right). The asterisks indicate the significance levels (``*'', ``**'', ``***'' for $p<0.05, 0.01, 0.001$ respectively). 
}
\label{fig:clustering}
\end{figure*}

EPLMs' fine-grained knowledge of moral norms, inspected in the previous experiment, might be more accuracte for western cultures than other cultures. We investigate this claim by clustering countries based on 1) their Western-Eastern economic status (i.e., Rich West grouping)\footnote{\url{https://worldpopulationreview.com/country-rankings/western-countries}}, and 2) their continent (i.e., geographical grouping). We repeat the experiments in the previous section for different country groups. The results are shown in Figure~\ref{fig:clustering}. We also try sampling the same number of countries in each group. The results remain robust and are illustrated in Appendix-\ref{sec:app_sample}. 

Our findings indicate that EPLMs contain more knowledge about moral norms of the Rich West countries as opposed to non-western and non-rich countries. Similarly, EPLMs have captured a more accurate estimation of the moral norms in countries located in Oceania, North America, and Europe, as opposed to African, Asian, and South American countries. The empirical moral norm ratings from European countries in WVS are highly aligned with North American countries ($r = 0.938$), which explains why their moral norms are inferred more accurately than non-English speaking countries.

Next, for each topic, we compare the z-scores of the empirical moral ratings with the z-scores of the \textsc{GPT3-PROBS} inferred moral scores, using Mann-Whitney U rank test. The results reveal that ``abortion’’, ``suicide’’, ``euthanasia’’, ``for a man to beat his wife’’, ``parents beating children’’, ``having casual sex’’, ``political violence’’, and ``death penalty’’ in non-western and non-rich countries are all encoded as more morally appropriate  than the actual data. Such misrepresentations of moral norms in these countries could lead to stereotypical content generation. We also find that For Rich West countries, ``homosexuality’’, ``divorce’’, and ``sex before marriage’’ are encoded as more morally inappropriate than the ground truth, ($p < 0.001$ for all, Bonferroni corrected). Such underlying moral biases, specifically toward ``homosexuality’’ might stimulate the generation of harmful content and stigmatization of members of LGBTQ+, which has been reported in BERT-based EPLMs~\cite{nozza-etal-2022-measuring}. The results for the rest of the models are similar and are shown in Table~\ref{tab:cluster_scores} in the Appendix.

Our method of clustering countries is simplistic and may overlook things such as the significant diversity in religious beliefs within the Non-Rich-West category, and thus it does not reflect the nuanced biases that models may possess when it comes to moral norms influenced by different religious traditions. Nonetheless, our approach still serves as a valuable starting point for studying EPLM's moral biases towards more fine-grained religious and ethnic communities.

\subsection{Cultural diversities and shared tendencies over the morality of different topics}
\label{sec:global_variation}
We next investigate whether EPLMs have captured the cultural diversities and shared tendencies over the morality of different topics (level 2 analysis). For example, people across cultures tend to disagree more about ``divorce'' than about ``violence against other people'' as depicted in Figure~\ref{fig:wvs_fig1}. Such cultural diversities for each topic can be measured by taking the standard deviation of the empirical moral ratings across different countries. The EPLMs' inferred cultural diversities can similarly be measured by taking the standard deviation of the estimated fine-grained moral scores for different countries. We then quantify the alignment between the two using Pearson correlation.

\begin{figure*}[ht!]
  \includegraphics[width=.99\linewidth]{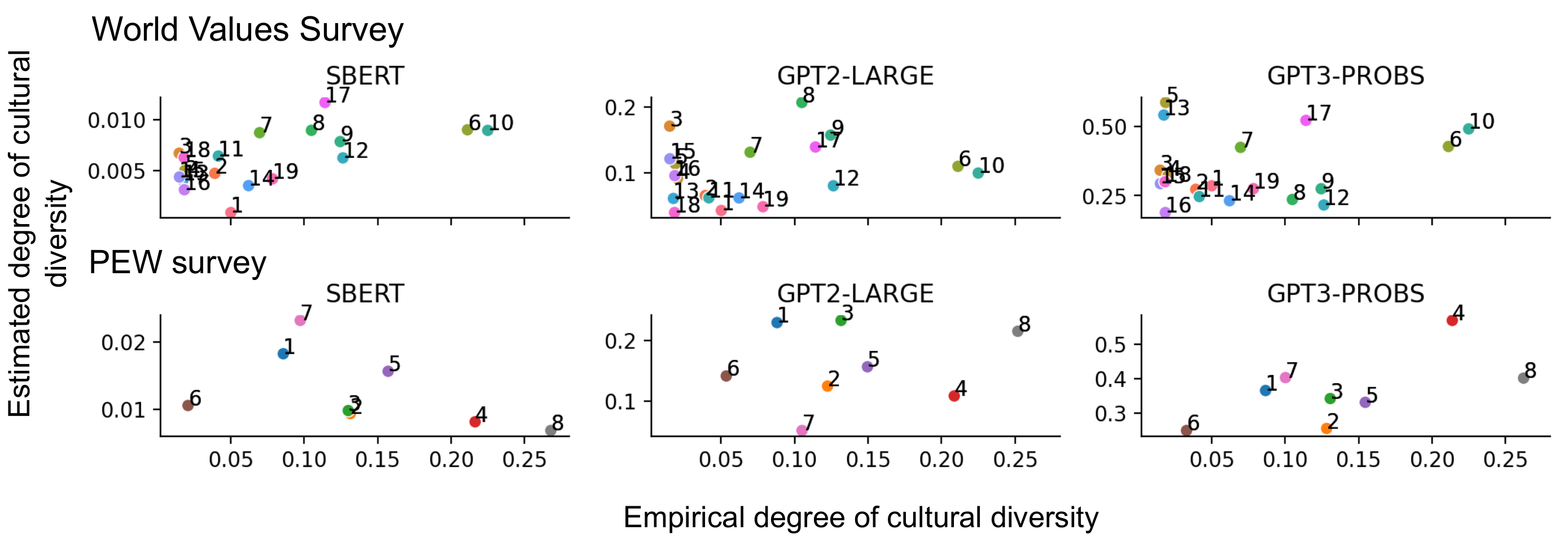}
  \caption{Comparison between the degrees of cultural diversities and shared tendencies in the empirical moral ratings and language-model inferred moral scores. Each dot corresponds to a moral topic. 
  The numerical indices are consistent with the legend indices in Table~\ref{tab:legends}.
  The x-axis shows the empirical standard deviations in moral ratings across countries and the y-axis shows the standard deviations from the model-inferred moral scores.
 }
  \label{fig:variation}
\end{figure*}

Figure~\ref{fig:variation} shows the results  for \textsc{SBERT, GPT2-LARGE, GPT3-PROBS}, and the rest are shown in Figure~\ref{fig:variation_app} in the Appendix. None of the correlations with the PEW survey were significant. For WVS, \textsc{SBERT, GPT2} and \textsc{GPT2-MEDIUM} exhibited a significant correlation ($p < 0.001$) with $r = 0.618$, $r = 0.579$, and $r = 0.734$ respectively. The results for \textsc{GPT3} are insignificant, suggesting that it is more challenging to correctly estimate cultural controversies of topics for \textsc{GPT3}. For example, \textit{stealing property} is incorrectly estimated to be more controversial than \textit{abortion}.

\section{Fine-tuning language models on global surveys}
Finally, we explore the utility-bias trade-off in encoding cultural moral knowledge into EPLMs by fine-tuning them on cross-cultural surveys. The utility comes from increasing the cultural moral knowledge in these models, and the bias denotes their decreased ability to infer English moral norms, in addition to the cultural moral biases introduced to the model. We run our experiments on \textsc{GPT2}, which our results suggest having captured minimum information about cultural moral norms compared to other autoregressive models.


\begin{table*}[ht]
\centering
\begin{center}
\scalebox{0.95}{
\begin{tabular}{ l| l| p{0.8in}p{0.9in} | p{0.8in}p{0.8in}}
 Train data& Data partition strategy & \multicolumn{2}{p{1.7in}|}{Evaluation}  &
\multicolumn{2}{p{1.6in}}
{\vspace{-0.2in}\makecell{\hspace{-0.4in}Performance on the \\ \hspace{-0.3in}Homogeneous norms }}\\ 
\hline
\multirow{3}{*}{WVS}
  & Random  &\textcolor{blue}{$\textbf{0.832}^{***}$}  $\uparrow$ & ($0.271^{***}$) & 
  \textcolor{red}{$0.71^{***}$} $\downarrow$ & \multirow{6}{*}{($0.80^{***}$)}
   \\
  & Country-based  & \textcolor{blue}{$0.759^{***}$} $\uparrow$ &($0.225^{**}$) &  \textcolor{red}{$0.72^{***}$} $\downarrow$ &
  \\
  & Topic-based  & \textcolor{blue}{$0.508^{***}$} $\uparrow$  &($0.286^{***}$) &   \textcolor{red}{$0.70^{***}$}  $\downarrow$ &\\
\cline{1-5}
\multirow{3}{*}{PEW} 
    & Random   & \textcolor{blue}{$\textbf{0.818}^{***}$}  $\uparrow$ & ($0.204$, n.s.) & \textcolor{red}{$0.64^{***}$}  $\downarrow$  & \\
  & Country-based  &  \textcolor{blue}{$0.764^{***}$}  $\uparrow$ & ($0.055$, n.s.) & \textcolor{red}{$0.67^{***}$}  $\downarrow$ & \\
  & Topic-based  &  \textcolor{blue}{$0.733^{***}$} $\uparrow$ &
  ($-0.146$, n.s.)
  & \textcolor{red}{$0.61^{***}$}  $\downarrow$  &\\
 \hline
\end{tabular}
}
\caption{Summary of fine-tuned \textsc{GPT2} language model performance on inferring moral norms across cultures and the degradation of its performance on inferring Homogeneous moral norms. Values in parentheses show the performance before fine-tuning. The arrows and colors show performance increase (blue, $\uparrow$) and decrease (red, $\downarrow$) after fine-tuning. 
 The asterisks indicate the significance levels (``*'', ``**'', ``***'' for $p<0.05, 0.01, 0.001$). 
}
\label{tab:gpt2_finetuning_mean}
\end{center}
\end{table*}

To fine-tune the model, for each participant from \texttt{[Country]} with \texttt{[Moral rating]} toward \texttt{[Topic]}, we designed a prompt with the structure ``\texttt{A person in [Country] believes [Topic] is [Moral rating]}.''. We used the surveys' wordings for \texttt{[Moral rating]}. Table~\ref{tab:finetuning_prompt} in the Appendix shows our prompts for WVS and PEW. These prompts constructed our data for fine-tuning, during which we maximize the probability of the next token. The fine-tuned models were evaluated on the same correlation tests introduced in the previous Sections~\ref{sec:cultural_variation},~\ref{sec:clusters}, and~\ref{sec:global_variation}. 

The fine-tuning data was partitioned into training and evaluation sets using different strategies (i.e., Random, Country-based, and Topic-based). For the Random strategy, we randomly selected $80\%$ of the fine-tuning data for training the model. The topic-country pairs not seen in the training data composed the evaluation set.  
For our Country-based and Topic-based strategies, we randomly removed $20\%$ of the countries ($n = 11$ for WVS, $n = 8$ for PEW) and topics ($n = 4$ for WVS, $n = 2$ for PEW) from the training data to compose the evaluation set. See Appendix~\ref{sec:app_finetuning} for the total number of samples.


Table~\ref{tab:gpt2_finetuning_mean} shows the gained utilities, that is the correlation test results between the fine-grained moral scores inferred by the fine-tuned models and the empirical fine-grained moral ratings. All fine-tuned models align better with the ground truth than the pre-trained-only models (i.e., the values in parentheses). For both WVS and PEW, the Random strategy is indeed the best as each country and topic are seen in the training data at least once (but may not appear together as a pair). The fine-tuned models can also generalize their moral scores to unseen countries and topics. Repeating the experiment in  Section~\ref{sec:global_variation} also shows substantial improvement in identifying cultural diversities of different topics by all fine-tuned models. For example, the WVS and PEW-trained models with Random strategy gain Pearson's r values of $0.893$, and $0.944$ respectively. The results for the rest of the models are shown in Table~\ref{tab:gpt2_finetuning_sd} in the Appendix. 

Nevertheless, the bias introduced during the fine-tuning decreases the performance on the Homogeneous norms dataset.
This observation displays a trade-off between cultural and homogeneous moral representations in language models. Moreover, injecting the cross-cultural surveys into EPLMs might introduce additional social biases to the model that are captured through these surveys~\cite{joseph-morgan-2020-word}. 

In addition, we probe the best fine-tuned model (i.e., WVS with Random strategy) on its ability to capture the moral norms of non-western cultures by repeating the experiment in Section~\ref{sec:clusters}. The results in Figure~\ref{fig:clustering} show that the fine-tuned \textsc{GPT2} performs the best for all country groups. There is still a gap between western and non-western countries. However, basic fine-tuning proves to 
be effective in adapting EPLMs to the ground truth.


\section{Discussion and conclusion}
We investigated whether English pre-trained language models contain knowledge about moral norms across many different cultures. Our analyses show that  large EPLMs  capture moral norm variation to a certain degree, with the inferred norms being predominantly more accurate in western cultures than non-western cultures. Our fine-tuning analysis further suggests that EPLMs' cultural moral knowledge can be improved using global surveys of moral norms, although this strategy reduces the capacity to estimate the English moral norms and potentially introduces new biases into the model. 
Given the increasing use of EPLMs in multicultural environments, our work highlights the importance of cultural diversity in automated inference of moral norms. Even when an action such as ``political violence’’ is assessed by an EPLM as morally inappropriate in a homogeneous setting, the same issue may be inferred as morally appropriate for underrepresented cultures in these large language models. Future work can explore alternative and richer representations of cultural moral norms that go beyond the point estimation we presented here and investigate how those representations might better capture culturally diverse moral views. 




\section*{Limitations}

Although our datasets are publicly available and gathered from participants in different countries, they cannot entirely represent the moral norms from all the individuals in different cultures over the world or predict how moral norms might change into the future~\cite{bloom2010morals, bicchieri2005grammar}. Additionally, we examine a limited set of moral issues for each country, therefore the current experiments should not be regarded as comprehensive  of the space of moral issues that people might encounter in different countries.

Moreover, taking the average of moral ratings for each culture is a limitation of our work and reduces the natural distribution of moral values in a culture to a single point~\cite{talat2021word}. Implementing a framework that incorporates both within-country variation and temporal moral variation~\cite{xie-etal-2019-text} is a potential future research direction.

Currently, it is not clear whether the difference between EPLMs’ estimated moral norms and the empirical moral ratings is due to the lack of cultural moral norms in the pre-training data, or that the cultural moral norms mentioned in the pre-training data represent the perspective of an English-speaking person of another country. For example, a person from the United States could write about the moral norms in another country from a western perspective. A person from a non-western country could also write about their own moral views using English. These two cases have different implications and introduce different moral biases into the system.

\section*{Potential risks}
We believe that the language models should not be used to prescribe ethics, and here we approach the moral norm inference problem from a descriptive perspective. However, we acknowledge modifying prompts could lead language models to generate ethical prescriptions for different cultures. Additionally, our fine-tuning approach could be exploited to implant cultural stereotypical biases into these models.


Many topics shown in this work might be sensitive to some people yet more tolerable to some other people. Throughout the paper, we tried to emphasize that none of the moral norms, coming from either the models' estimation or the empirical data, should be regarded as definitive values of right and wrong, and the moral judgments analyzed in this work do not reflect the opinions of the authors.




\section*{Acknowledgements}
This work was supported by a SSHRC Insight Grant 435190272.

\clearpage
\appendix
\bibliography{anthology,custom,biblio}
\bibliographystyle{acl_natbib}

\appendix

\label{sec:appendix}

\section{Data license}
Both World Values Survey and PEW survey are publicly available to use for research purposes. We accept and follow the terms and conditions for using these datasets, which can be found in \url{https://www.worldvaluessurvey.org/WVSContents.jsp?CMSID=Documentation}, and \url{https://www.pewresearch.org/about/terms-and-conditions/}.

\section{Comparison of human-rated and machine-scored moral norms}
\begin{figure*}[!htbp]
  \includegraphics[width=\textwidth]{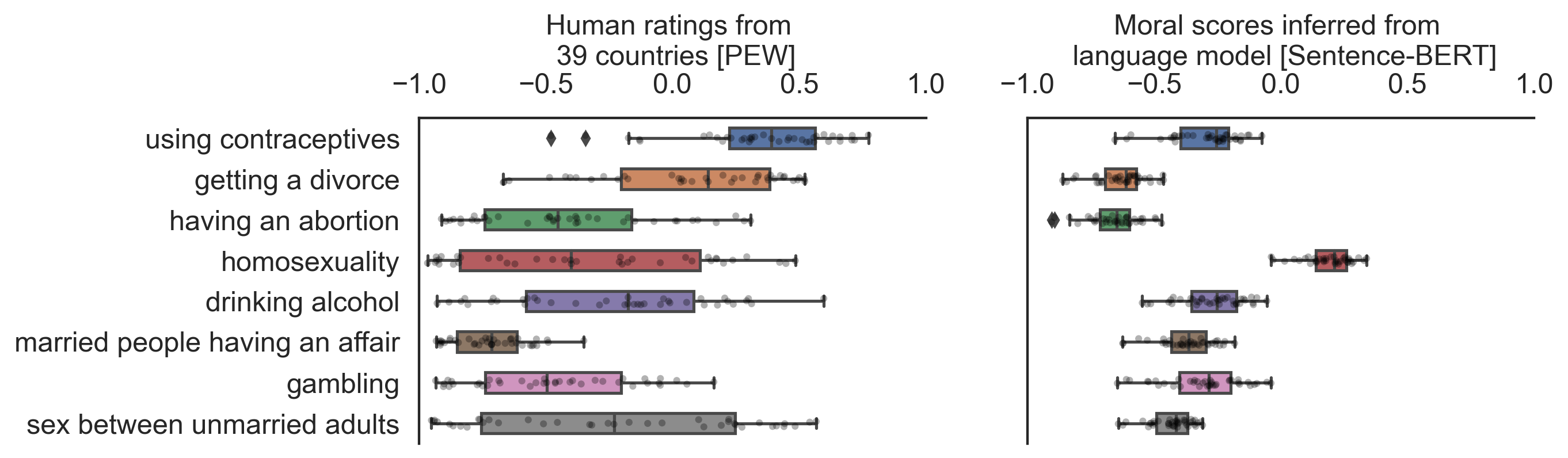}
  \caption{Comparison of human-rated and machine-scored moral norms across cultures. Left: Boxplots of human ratings of moral norms across cultures in the PEW survey. Each dot represents the empirical average of participants' ratings for a morally relevant topic (e.g., ``having an abortion'') within a country. Right: Corresponding moral scores estimated by a language model (Sentence-BERT)~\cite{reimers-2019-sentence-bert}. Each dot represents the moral score obtained by probing the language model in a given country.}
  \label{fig:pew_fig1}
\end{figure*}

Figure~\ref{fig:pew_fig1} shows the comparison between human-rated moral norms in PEW, and the moral scores inferred by SBERT~\cite{reimers-2019-sentence-bert}.

\section{Probing experiments}
\label{sec:prompts_details}

Table~\ref{tab:prompt} shows our prompt design for probing fine-grained moral norms in EPLMs. As mentioned in the main text, we  repeat our probing experiment for \textsc{GPT2} models and \textsc{GPT3-PROBS} with another template ``People in [Country] believe [Topic] is [Moral Judgment]''. The results are substantially worse than our initial template, suggesting that extracting the moral knowledge in language models is sensitive to the wording used in the input. The results for the fine-grained analysis (level 1 analysis) and the cultural diversities and shared tendencies (level 2 analysis) with this template are shown in Table~\ref{tab:new_prompts}.

In all experiments, we used a single NVIDIA TITAN V GPU. Each probing experiment took approximately 1 hour to complete. 

\begin{table*}[htbp!]
\begin{center}
\begin{tabular}{l|l} 
Model & Prompt\\
\hline
 Sentence-BERT &\texttt{[Topic] in [Country].}  \\
 \hline
\multirow{2}{*}{GPT2 models and GPT3-PROBS} & \texttt{In [Country] [Topic] is [Moral judgement].} \\
\cline{2-2}
& \texttt{People in [Country] believe [Topic] is [Moral judgement].} \\
\hline
GPT3-QA (for PEW) & \makecell[l]{\texttt{Do people in [Country] believe that [Topic] is: }\\ \texttt{1) Morally acceptable} \\ \texttt{2) Not a moral issue} \\ \texttt{3) Morally unacceptable.}} \\
\hline
GPT3-QA (for WVS) & \makecell[l]{\texttt{Do people in [Country] believe that [Topic] is: }\\ \texttt{1) Always Justifiable} \\ \texttt{2) Something in between} \\ \texttt{3) Never justifiable.}} \\
\end{tabular}
\caption{Prompting design used for estimating the fine-grained moral norms in different language models. In our homogeneous norm inference, we remove ``\texttt{In [country]}'' from the prompts.}
\label{tab:prompt}
\end{center}
\end{table*}

\section{Homogeneous moral norm inference}
Table~\ref{tab:varrying_corr} shows the detailed values of the correlation tests in our homogeneous moral norm inference experiment. 
\begin{table*}[!htbp]
\begin{center}
\scalebox{0.875}{
\begin{tabular}{ l|c |c | c} 
Model & \makecell{World Values Survey} ($n = 1,028$)& PEW survey ($n = 312$) & Homogeneous norms ($n = 100$)\\
\hline
 SBERT &  $0.21 0^ {***}$ & $-0.038$ (n.s.) & $0.79^ {***}$\\
GPT2 &  $0.176^ {***}$ & $-0.069$ (n.s.) & $0.80^ {***}$ \\
GPT2-MEDIUM &  $0.181^ {***}$ & $0.033$ (n.s.) & $0.79^ {***}$ \\
GPT2-LARGE &  $0.226^ {***}$ & $0.157$ (n.s.)  & $0.76^ {***}$  \\
GPT3-QA & $0.330^ {***}$ & $0.391^ {***}$ & $0.79 ^ {***}$\\
GPT3-PROBS  & $0.346^ {***}$ & $0.340^ {***}$ & $0.85 ^ {***}$ \\
\end{tabular}
}
\caption{
Performance of pre-trained language models (without cultural prompts) on inferring 1) homogeneous westernized moral norms, and 2) culturally diverse moral norms recorded in World Values Survey and PEW survey data.}
\label{tab:varrying_corr}
\end{center}
\end{table*}

\begin{table*}[h!]
\begin{center}
\begin{tabular}{  l|l|c|c} 
Data & model &  \makecell{Fine-grained evaluation \\of moral norms} & \makecell{Evaluation on cultural diversity \\ and shared tendencies}\\ 
\hline
\multirow{4}{*}{WVS} & \textsc{GPT3-PROBS} & \hspace{-2.5mm}$0.078^{*}$ & $-0.176$ \\
& \textsc{GPT2} & $-0.114^{***}$ & \hspace{2.5mm}$0.231$\\
& \textsc{GPT2-MEDIUM} & $-0.261^{***}$ & $-0.357$ \\
& \textsc{GPT2-LARGE} & \hspace{-4.5mm}$-0.07^{*}$ & $-0.356$ \\
  \hline
  \multirow{4}{*}{PEW} & \textsc{GPT3-PROBS} & \hspace{2.75mm}$0.539^{***}$ & \hspace{2.5mm}$0.041$ \\
  & \textsc{GPT2} & \hspace{0.5mm} $0.168^{**}$ & \hspace{1.5mm} $0.566$ \\
  & \textsc{GPT2-MEDIUM} & \hspace{1.5mm}$0.165^{**}$ & \hspace{1.5mm} $0.184$ \\
  & \textsc{GPT2-LARGE} & \hspace{1mm}$0.19^{***}$ & \hspace{1.5mm} $0.542$ \\
\end{tabular}
\caption{Performance of pre-trained autoregressive language models on identifying fine-grained moral norms and cultural diversities and shared tendencies, using the prompt template ``People in [Country] believe [Topic] is [Moral judgment]". The values are Pearson's correlations. The asterisks indicate the significance levels (``*'', ``**'', ``***'' for $p<0.05, 0.01, 0.001$ respectively). }
\label{tab:new_prompts}
\end{center}
\end{table*} 

\section{Fine-grained cultural variation of moral norm}
\begin{figure*}[!htbp]
\includegraphics[width=1\textwidth]{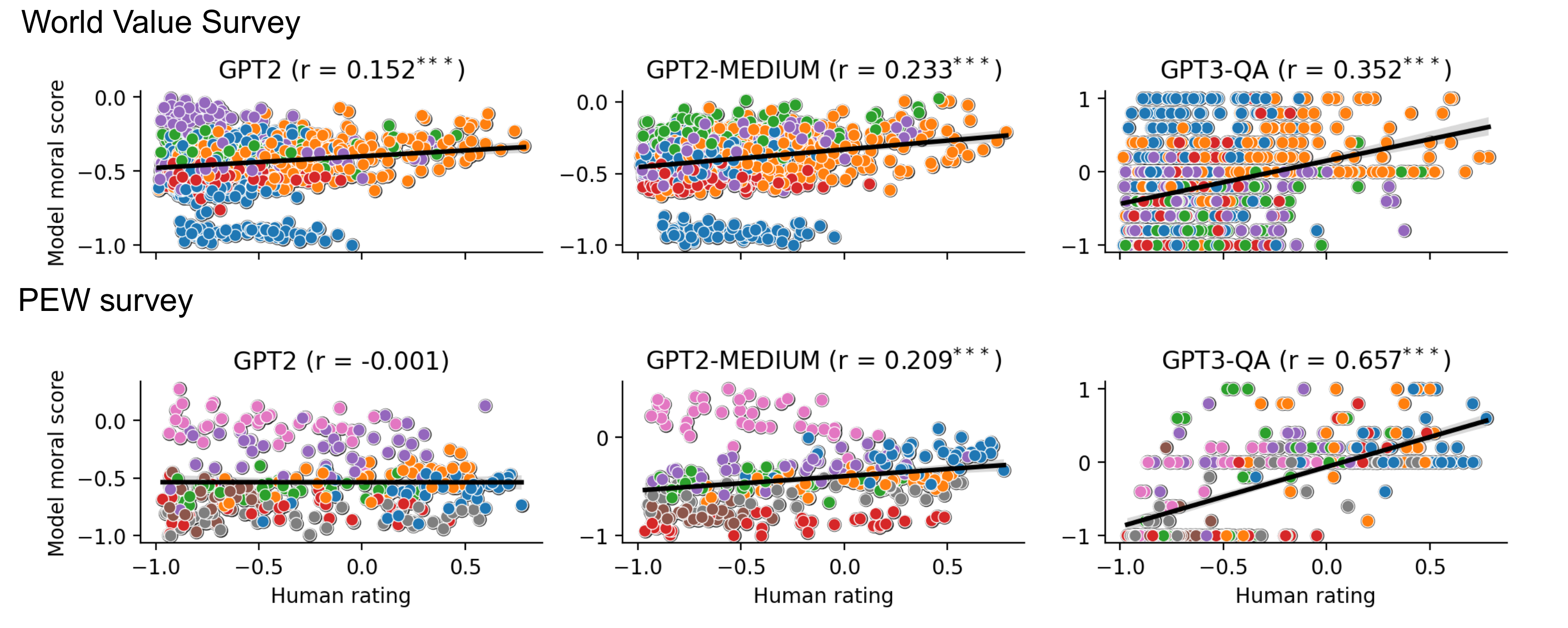}
\caption{Degree of alignment between the moral scores from EPLMs and fine-grained empirical moral ratings for different topics across countries taken from the World Values Survey (top) and PEW survey (bottom). 
Each dot represents a topic-country pair. The x-axis shows the fine-grained moral ratings from the ground truth and the y-axis shows the corresponding inferred moral scores. Similar topics in the World Value Surveys are shown with the same color.
}
  \label{fig:figure2_app}
\end{figure*}

\begin{figure*}[!htbp]
  \includegraphics[width=\linewidth]{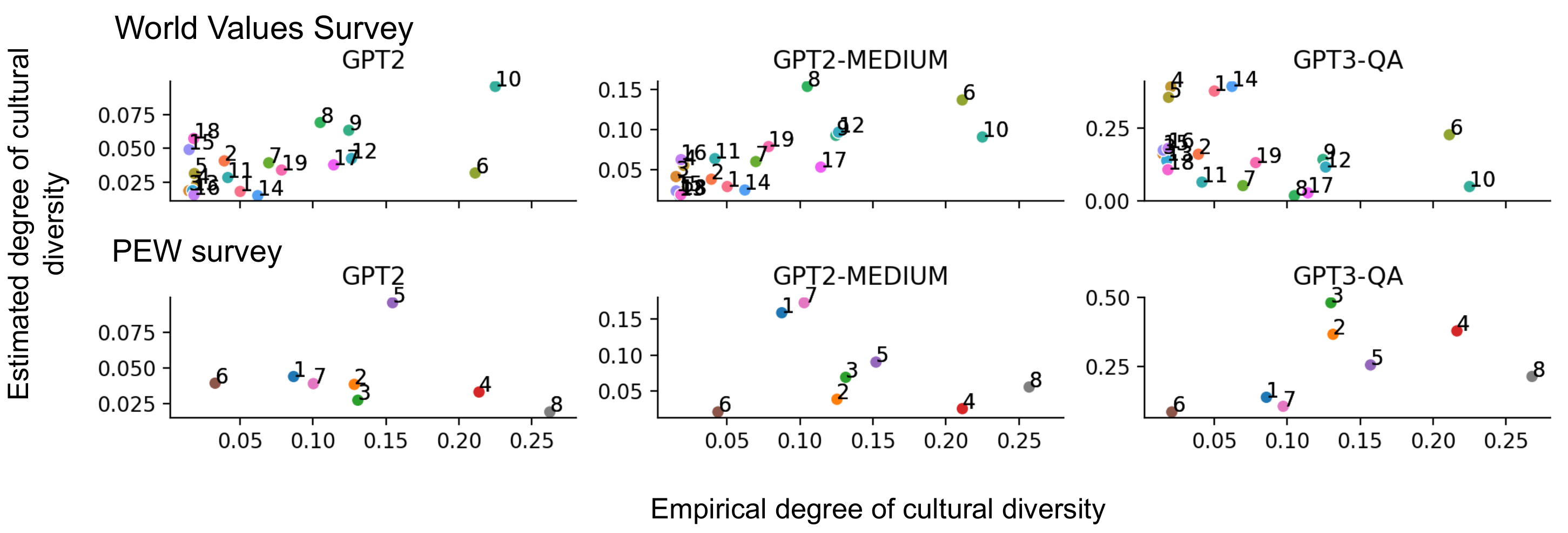}
  \caption{Comparison between the degrees of cultural diversities and shared tendencies in the empirical moral ratings and language-model inferred moral scores. Each dot corresponds to a moral topic. The x-axis shows the empirical standard deviations in moral ratings across countries and the y-axis shows the standard deviations from the model-inferred moral scores.}
  \label{fig:variation_app}
\end{figure*}

Figure~\ref{fig:figure2_app} and Figure~\ref{fig:variation_app} show the result of our fine-grained cultural moral inference, and inference of cultural diversities and shared tendencies respectively
for \textsc{GPT2, GPT2-MEDIUM}, and \textsc{GPT3-QA}. The numerical indices in Figure~\ref{fig:variation_app} are consistent with the indices in Table~\ref{tab:legends}.

\begin{table*}[]
    \centering
    \begin{tabular}{l|l}
    \multicolumn{2}{l}{World Values Survey} \\ \hline
         Index & Topic \\ \hline
         1& stealing property \\
         2 & euthanasia \\
         3 & sex before marriage \\
         4 & violence against other people\\
         5 & cheating on taxes \\
         6 & avoiding a fare on public transport \\
         7 & abortion \\
         8 & suicide \\
         9 & someone accepting a bribe on a course of their duties \\
         10 & terrorism as a political, ideological, or religious mean\\
         11 & homosexuality\\
         12 & parents beating children\\
         13 & prostitution \\
         14 & divorce \\
         15 & political violence \\
         16 & death penalty \\
         17 & claiming governments benefits to which you are not entitled \\
         18 & for a man to beat his wife\\
         19 & having casual sex \\ \hline
         \multicolumn{2}{l}{PEW survey} \\ \hline
         1 & using contraceptives \\
         2 & getting a divorce \\
         3 & having an abortion \\
         4 & homosexuality \\
         5 & drinking alcohol\\
         6 & married people having an affair \\
         7 & gambling \\
         8 & sex between unmarried adults
    \end{tabular}
    \caption{Numerical indexing for topics in moral surveys.}
    \label{tab:legends}
\end{table*}

\section{Sampling for cultural clusters}
\label{sec:app_sample}

\begin{figure*}
\includegraphics[width=\textwidth]{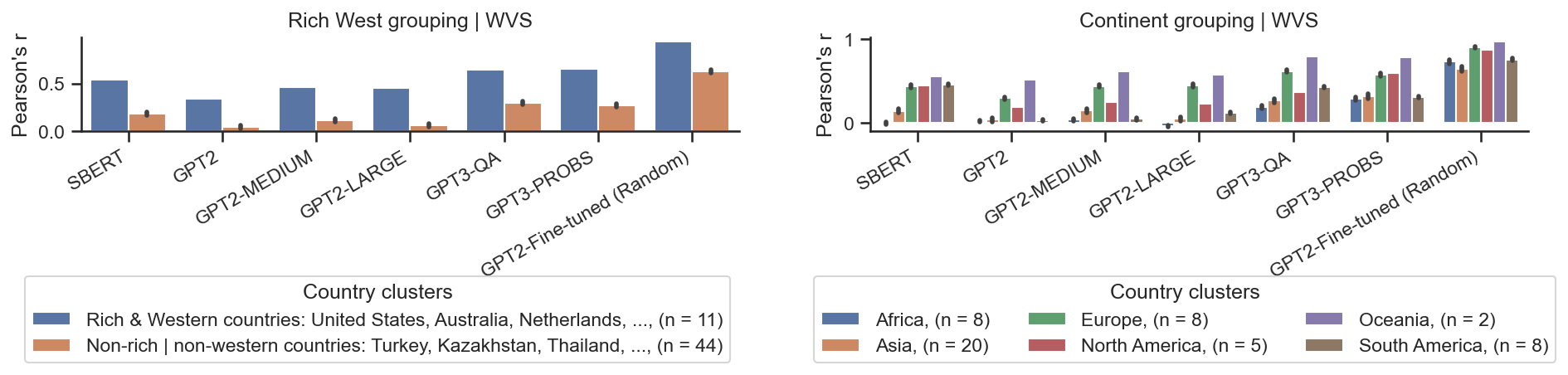}
\caption{Correlation between language-model inferred moral scores and empirical moral ratings from World Values Survey analyzed in different clusters of countries in Rich West grouping (left) and continent grouping (right). The results are generated by sampling and the error bars show the confidence intervals with $\alpha = 0.05$.
}
\label{fig:clustering_2}
\end{figure*}

Since in section~\ref{sec:clusters} there are a different number of countries in each group, we redo the experiment by randomly sampling the same number of countries ($n = 11$ for Rich West grouping, $n = 5$ for continent grouping) and repeating the sampling process for $50$ times. The results and the general pattern remain the same and are depicted in Figure~\ref{fig:clustering_2}.

\begin{table*}
\scalebox{0.9}{
\begin{tabular}{l | l | l}
Model &  \makecell[l]{Positively evaluated topics \\for non-rich and non-western countries} & \makecell[l]{Negatively evaluated topics \\for Rich-West countries} \\
\hline
\textsc{SBERT} & \makecell[l]{
sex before marriage$^{**}$, homosexuality$^{***}$, \\having casual sex$^{***}$, abortion$^{***}$,\\ prostitution$^{***}$, claiming government\\ benefits to which you are\\ not entitled$^{***}$, someone \\accepting a bribe \\in the course of their duties$^{***}$}
& \makecell[l]{
sex before marriage$^{***}$, euthanasia$^{***}$,\\ divorce$^{***}$, death penalty$^{***}$, \\parents beating children$^{***}$}
\\
\hline
\textsc{GPT2} & 
\makecell[l]{
abortion$^{***}$, prostitution$^{***}$,\\ suicide$^{***}$, avoiding a fare on \\public transport$^{***}$,\\ someone accepting a bribe\\ in the course of their duties$^{***}$,\\ terrorism as a political,\\ ideological or religious mean$^{***}$, \\political violence$^{***}$, \\violence against other people$^{***}$
}
& \makecell[l]{
sex before marriage$^{**}$, homosexuality$^{**}$,\\ divorce$^{**}$, having casual sex$^{**}$,\\ claiming government benefits \\to which you are not entitled$^{***}$}
\\ \hline
\textsc{GPT2-MEDIUM} & 
\makecell[l]{
euthanasia$^{***}$, abortion$^{***}$, suicide$^{***}$, \\avoiding a fare on public transport$^{***}$, \\someone accepting a bribe in \\the course of their duties$^{***}$,\\ political violence$^{***}$, violence against\\ other people$^{***}$, stealing property$^{***}$
} & \makecell[l]{
sex before marriage$^{***}$, homosexuality$^{**}$,\\ divorce$^{**}$, having casual sex$^{**}$,\\ claiming government benefits \\to which you are not entitled$^{***}$
}
\\
\hline
\textsc{GPT2-LARGE} & \makecell[l]{
euthanasia$^{***}$, having casual sex$^{***}$, \\abortion$^{***}$, prostitution$^{***}$, suicide$^{***}$,\\ terrorism as a political, \\ideological or religious mean$^{***}$, \\political violence$^{***}$, \\violence against other people$^{***}$} & \makecell[l]{sex before marriage$^{***}$, homosexuality$^{**}$,\\
divorce$^{**}$, claiming government benefits to \\which you are not entitled$^{***}$}
\\ \hline
\textsc{GPT3-QA} & \makecell[l]{
     having casual sex$^{**}$, abortion$^{**}$, \\avoiding a fare on public transport$^{***}$, \\cheating on taxes$^{***}$, \\someone accepting a bribe in the\\ course of their duties$^{***}$, \\political violence$^{***}$
     }
     & \makecell[l]{sex before marriage$^{***}$, divorce$^{**}$, \\death penalty$^{**}$, prostitution$^{**}$,\\ parents beating children$^{**}$,\\ suicide$^{**}$, for a man to beat his wife$^{***}$,\\ stealing property$^{**}$
     }
     \\ \hline
     \textsc{GPT3-PROBS} &
     \makecell[l]{euthanasia$^{***}$, having casual sex$^{***}$,\\ abortion$^{***}$, death penalty$^{***}$, \\suicide$^{***}$, political violence$^{***}$, \\for a man to beat his wife$^{***}$}
     &\makecell[l]{sex before marriage$^{***}$, homosexuality$^{***}$ ,\\ divorce$^{**}$}
     \\ 
    
\end{tabular}
}
\caption{Topics evaluated as morally positive for non-rich and non-western countries and morally negative for Rich-West countries, in comparison to the ground truth in these countries. In each entry, the topics are sorted from the most controversial (i.e., having the highest degree of cultural diversity) to the least controversial. The asterisks indicate the significance levels of Mann-Whitney U rank test after Bonferroni p-value correction (``*'', ``**'', ``***'' for $p<0.05, 0.01, 0.001$ respectively).}
\label{tab:cluster_scores}
\end{table*}

\section{Details of fine-tuning on global surveys}
\label{sec:app_finetuning}
Table~\ref{tab:finetuning_prompt} shows 
the \texttt{Moral rating} in our prompt design for constructing our fine-tuning dataset. For example, The World Value Survey represents the two ends of the ratings scale where 1 is ``Never justifiable'' and 10 is ``Always justifiable''. The options in between are presented to the participants in a 10-point scale. Therefore, we mapped these options to different prompts that are semantically similar and in between the two ends. For example, if a participant from the United States rated \textit{stealing property} as 2, which is slightly more positive than the first option (``Never justifiable''), we mapped this rating to ``not justifiable'', creating the prompt ``\texttt{A person in the United States believes stealing property is not justifiable.}'' for our fine-tuning data.

Since there are a different number of participants from each country, in order to balance this dataset, we randomly select $100$ samples for each topic-country pair and removed the rest of the utterances from the training data. We fine-tuned \textsc{GPT2} on one epoch, with a batch size of $8$, learning rate of $5e-5$, and weight decay of $0.01$. The number of training and evaluation samples for all data partition strategies are shown in Table~\ref{tab:finetuning_meta}. In all experiments, we used a single NVIDIA TITAN V GPU.
Fine-tuning and evaluation took approximately 2 hours to complete for each model.

\begin{table*}[h!]
\begin{center}
\begin{tabular}{  l|l |p{0.7in}p{0.7in}} 
 Train data &  Data partition strategy &\multicolumn{2}{p{1.4in}}{  \makecell{Evaluation}}\\ 
\hline
\multirow{3}{*}{WVS} 
  & Random  &{\textcolor{blue}{$0.893^{***}$}}  $\uparrow$ & \multirow{3}{*}{($0.579^{***}$) } \\
  & Country-based  & \textcolor{blue}{$0.894^{***}$}  $\uparrow$ & \\
  & Topic-based  & \textcolor{blue}{$0.835^{***}$} $\uparrow$ & \\
  \hline
\multirow{3}{*}{PEW} 
  & Random  & \textcolor{blue}{$0.944^{**}$}  $\uparrow$ &\multirow{3}{*}{(n.s.)} \\
  & Country-based  &  {\textcolor{blue}{$0.839 ^{*}$} }$\uparrow$&  \\
  & Topic-based   &\textcolor{blue}{$0.953^{***}$}  $\uparrow$ &\\ \hline
\end{tabular}
\caption{Summary of fine-tuned \textsc{GPT2} language model performance in inferring the cultural diversities and shared tendencies over the morality of different topics. The arrows and colors show performance increase (blue, $\uparrow$) and decrease (red, $\downarrow$) after fine-tuning. All values are Pearson's correlations. The asterisks indicate the significance levels (``*'', ``**'', ``***'' for $p<0.05, 0.01, 0.001$ respectively). 
Non-significant results are shown by ``n.s.''. }
\label{tab:gpt2_finetuning_sd}
\end{center}
\end{table*} 

\begin{table*}[!htbp]
\begin{center}
\begin{tabular}{|l|c |c |} 
\hline
Dataset & Rating & \texttt{[Moral rating]} in fine-tuning prompts \\
\hline
\multirow{5}{*}{WVS} & 1 & \texttt{never justifiable} \\
\cline{2-3}
& [2, 3, 4]& \texttt{not justifiable} \\ \cline{2-3}
& [5, 6] & \texttt{somewhat justifiable} \\\cline{2-3}
& [7, 8, 9] & \texttt{justifiable} \\
\cline{2-3}
& 10 & \texttt{always justifiable} \\
\hline 
\multirow{3}{*}{PEW} & 1 & \texttt{morally unacceptable} \\ \cline{2-3}
 & 2 & \texttt{not a moral issue} \\ \cline{2-3}
 & 3 & \texttt{morally acceptable} \\ \hline
\end{tabular}
\caption{Different prompting designs for fine-tuning language models on the global survey datasets.}
\label{tab:finetuning_prompt}
\end{center}
\end{table*}

\begin{table*}[!htbp]
\begin{center}
    \begin{tabular}{l|c|c|c}
        Data &  Data partition strategy & Training samples &
         Evaluation sample pairs
         \\
         \hline
          \multirow{3}{*}{WVS} & Random & 82200  
          & 206\\
          & Country-based & 82600   & 202 \\
         & Topic-based & 81200    & 216 \\
         
        \hline
          \multirow{3}{*}{PEW} & Random & 24900  & 63\\
         &  Country-based & 24800   & 64 \\
         & Topic-based  & 23400    & 78 \\
    \end{tabular}
    \caption{Number of samples in training and evaluation datasets for fine-tuning \textsc{GPT2} on global surveys of morality.}
\label{tab:finetuning_meta}
\end{center}
    
\end{table*}


\end{document}